\def\year{2022}\relax
\documentclass[letterpaper]{article} % DO NOT CHANGE THIS
\usepackage{aaai22}  % DO NOT CHANGE THIS
\usepackage{times}  % DO NOT CHANGE THIS
\usepackage{helvet}  % DO NOT CHANGE THIS
\usepackage{courier}  % DO NOT CHANGE THIS
\usepackage[hyphens]{url}  % DO NOT CHANGE THIS
\usepackage{graphicx} % DO NOT CHANGE THIS
\def\UrlFont{\rm}  % DO NOT CHANGE THIS
\usepackage{natbib}  % DO NOT CHANGE THIS AND DO NOT ADD ANY OPTIONS TO IT
\usepackage{caption} % DO NOT CHANGE THIS AND DO NOT ADD ANY OPTIONS TO IT
\DeclareCaptionStyle{ruled}{labelfont=normalfont,labelsep=colon,strut=off} % DO NOT CHANGE THIS
\usepackage{algorithm}
\usepackage{algorithmic}
\usepackage{subcaption}
\usepackage{amsfonts}
\usepackage{amsmath}
\usepackage[table]{xcolor}
\usepackage{tcolorbox}

\usepackage{multirow}
\usepackage{graphicx}
\usepackage{lipsum}
\usepackage{booktabs}
\usepackage{newfloat}
\usepackage{listings}
\floatstyle{ruled}
\newfloat{listing}{tb}{lst}{}
\floatname{listing}{Listing}
\title{Predicting and Understanding Human Action Decisions: Insights from Large Language Models and Cognitive Instance-Based Learning}

\author{
    Thuy Ngoc Nguyen\textsuperscript{\rm 1}, 
    Kasturi Jamale\textsuperscript{\rm 1}, 
    Cleotilde Gonzalez\textsuperscript{\rm 2}\\
}
\affiliations{
    \textsuperscript{\rm 1} Department of Computer Science, University of Dayton, OH, 45469, USA\\
    \textsuperscript{\rm 2} Department of Social and Decision Sciences, Carnegie Mellon University, PA, 15213, USA\\
    ngoc.nguyen@udayton.edu, jamalek1@udayton.edu, coty@cmu.edu
}

\usepackage{bibentry}

\begin{document}

\maketitle

\begin{abstract}

% Large Language Models (LLMs) have demonstrated their capabilities across a range of tasks, from language translation to complex reasoning. Yet, the ability of these models to understand and predict human behavior and biases is still an open question. This paper addresses this gap by leveraging the reasoning and generative capabilities of LLMs to predict human behavior in two sequential decision-making tasks. These tasks involve balancing between exploitative and exploratory actions and handling delayed feedback—both critical for simulating real-life decision processes. We compare the performance of LLMs with a cognitive instance-based learning (IBL) model, which simulates human experiential decision-making. Our findings indicate that LLMs excel at rapidly incorporating feedback to enhance prediction accuracy. In contrast, the IBL model better accounts for human exploratory behaviors and effectively captures loss aversion bias, even with limited experience. The study highlights the benefits of integrating LLMs with cognitive architectures, suggesting that this synergy could enhance the modeling and understanding of complex human decision-making patterns.

Large Language Models (LLMs) have demonstrated their capabilities across various tasks, from language translation to complex reasoning. Understanding and predicting human behavior and biases are crucial for artificial intelligence (AI)-assisted systems to provide useful assistance, yet it remains an open question whether these models can achieve this. This paper addresses this gap by leveraging the reasoning and generative capabilities of the LLMs to predict human behavior in two sequential decision-making tasks. These tasks involve balancing between exploitative and exploratory actions and handling delayed feedback—both essential for simulating real-life decision processes. We compare the performance of LLMs with a cognitive instance-based learning (IBL) model, which imitates human experiential decision-making. Our findings indicate that LLMs excel at rapidly incorporating feedback to enhance prediction accuracy. In contrast, the cognitive IBL model better accounts for human exploratory behaviors and effectively captures loss aversion bias — the tendency to choose a sub-optimal goal with fewer step-cost penalties rather than exploring to find the optimal choice, even with limited experience. The results highlight the benefits of integrating LLMs with cognitive architectures, suggesting that this synergy could enhance the modeling and understanding of complex human decision-making patterns.

\end{abstract}

\section{Introduction}
\label{sec:intro}
%#1: The importance of AI in understanding human behavior, particularly in decision-making
% With the rapid growth of artificial intelligence (AI) technologies in recent years, the integration of AI assistance into decision-making processes has become increasingly ubiquitous in many aspects of people's lives, become part of human-AI teaming. 
Understanding and predicting human behavior in decision-making settings is crucial for developing AI systems that can effectively collaborate with and assist people to help them make informed decisions and avoid cognitive biases and limitations~\cite{hoffman2023inferring, bansal2019updates, zhang2021ideal, rastogi2023taxonomy}.
One common approach to predicting individual behavior is using machine learning techniques to model their decision-making processes based on past behaviors. These techniques include imitation learning (e.g., behavior cloning~\cite{torabi2018behavioral}) and machine theory of mind (e.g., inverse reinforcement learning (RL)~\cite{abbeel2004apprenticeship}, Bayesian Theory of Mind~\cite{baker2017rational}, or neural networks~\cite{rabinowitz2018machine}). However, these methods often require extensive training datasets and struggle to model human decision-making accurately with limited samples.

% #2: Research Gap :
% Identify the specific gap in the current research landscape, particularly focusing on whether LLMs can predict and understand human decision-making effectively.
% Mention the role of cognitive models, like instance-based learning (IBL), in studying human behavior, and highlight the need to explore how these models compare to LLMs in simulating decision-making processes.

The recent rise of large language models (LLMs) such as ChatGPT~\cite{ouyang2022training}, PaLM~\cite{chowdhery2023palm}, and LLaMA~\cite{touvron2023llama} has demonstrated their remarkable capabilities in semantic understanding and intent reasoning~\cite{brown2020language,zhang2023recommendation}
by encoding a wide range of human behaviors from their training data. These advancements offer new opportunities for employing LLMs as work assistants, particularly in creating LLM-powered decision support systems~\cite{allen2023power,chiang2024enhancing,wang2024lave}.

A growing body of research has shown that these LLMs can perform at human levels, or even above, in many  experiments~\cite{binz2023using,dasgupta2022language,shiffrin2023probing} and tasks designed to test different aspects of reasoning~\cite{mahowald2024dissociating}. However, empirical findings on their ability to reason about the mental states of others, known as theory of mind, are mixed. While some studies show promising results~\cite{strachan2024testing}, others highlight limitations in accurately reasoning about the mental states of others in different theory of mind tasks~\cite{ullman2023large}. Furthermore, from an empirical standpoint, little is currently understood about whether these models can predict and capture human-like behavioral characteristics, especially human cognitive biases~\cite{mitchell2023debate}. For instance, an experiment in sequential decision-making that required a trade-off between exploitation and exploration showed that GPT-3 outperformed human subjects by heavily relying on exploitative strategies~\cite{binz2023using}. In contrast, people tended to apply a combination of elaborate exploration strategies~\cite{wilson2014humans}.

% Cognitive architectures play a crucial role in studying human behavior by providing an integrated view of the cognitive abilities of the human mind~\cite{anderson2014atomic}. 
Prior research has shown that humans rely on various cognitive mechanisms when making decisions~\cite{gonzalez2003instance,erev2010choice,lebiere2013functional}. These cognitive models have been instrumental in understanding the strengths and limitations of human performance and machine learning algorithms~\cite{thomson2014human,mitsopoulos2022toward}. With the rise of LLMs, how these cognitive models compare to LLMs in predicting human decision-making strategies is unclear. Addressing this gap is crucial for gaining deeper insights into LLMs' potential, providing a cognitive grounding between human users and these models, and guiding the development of LLM systems that can effectively interact with people.

% #3: Objective and Scope of the Paper
In this work, we investigate the capabilities of LLMs, specifically open-source models, in predicting human action strategies in two sequential decision-making tasks, and compare their performance with a cognitive instance-based learning (IBL) model~\cite{gonzalez2003instance}. Grounded in the theory of decisions from experience, IBL models simulate human decision-making by incorporating mechanisms and limitations from the ACT-R cognitive architecture~\cite{anderson2014atomic}. These models have proven effective in emulating human decisions in various tasks, including gambling choices~\cite{gonzalez2011instance,hertwig2015decisions}, complex dynamic resource allocation~\cite{somers2020cognitive}, cybersecurity~\cite{gonzalez2020design}, and predicting the actions of other RL agents~\cite{nguyen2022theory}

% #4: Methodology Preview (1 paragraph): 2 interactive tasks
Our goal is to understand whether LLMs and the cognitive IBL model can predict human action strategies and capture human biases, such as loss aversion, characterized by the tendency to choose sub-optimal goals with fewer step-cost penalties rather than exploring optimal choices. We focus on multi-step, goal-directed decision-making tasks in interactive environments that require balancing exploitative and exploratory actions and handling delayed feedback—essential components of real-life decision processes. 

To achieve this, we analyze the discrepancies between the strategy predictions of the models and real human strategies, which enabled us to uncover the ability of these models to capture the nuances of human behavior in balancing risk and reward during decision-making. We used schema-based and demonstration-based prompts to provide task instructions and users' action trajectory history from previous trials, allowing pre-trained LLMs to use this in-context information to predict the next action plans in subsequent trials. We employed two open-source LLMs for our experiments: Mistral 7B~\cite{jiang2023mistral} and Llama-3 70B (the largest of Meta AI's Llama-3 models with 8B and 70B parameters)~\cite{touvron2023llama,meta_llama_3}. We chose these state-of-the-art open-source LLMs over a closed-source commercial service like GPT-4~\cite{achiam2023gpt} as they provide transparency and public access, thus promoting reproducibility and responsible LLM use by giving researchers full access to the network architecture and its pre-trained weights.

% #5: Significance of the Research (1-2 paragraphs):
% The results from comparing the predicted behaviors of the model and humans, LLMs demonstrated a superior ability to rapidly incorporate feedback and improve prediction accuracy. In contrast, the IBL model more accurately accounted for human exploratory behaviors and effectively captured the loss aversion bias—the tendency to choose sub-optimal goals with fewer step-cost penalties. These findings suggest that integrating LLMs with cognitive architectures could enhance the modeling and understanding of complex human decision-making patterns. 

Our results from comparing the predicted behaviors of the models and humans demonstrate that the lightweight Mistral-7B model outperforms both Llama-3 70B and the cognitive IBL model in predicting human strategies. The LLMs also demonstrated an ability to quickly incorporate feedback and improve prediction accuracy as more demonstrated data was provided. As expected, predicting human behavior is more challenging in complex decision environments with high cost-reward tension. Importantly, we observed that the cognitive IBL model more accurately accounted for human exploratory behavior with few samples and aligned closely with human exploratory strategies under limited information, which reflects the tendency towards risk-averse or ``satisficing'' behavior~\cite{simon1956rational} to choose the closest sub-optimal option instead of seeking the optimal one. These findings suggest that integrating LLMs with cognitive architectures could enhance the modeling and understanding of complex human decision-making patterns.

% With the remarkable achievements of LLMs, these models are now at the forefront of integrating AI systems into everyday human interactions, proving beneficial for real-world applications involving human-in-the-loop control~\cite{allen2023power, zhao2023survey}. Understanding their ability to infer mental states from observed actions, similar to theory of mind abilities, is crucial. Research in this area yields mixed results, with some studies suggesting theory of mind-like abilities~\cite{kosinski2023theory}, while others find these abilities less clear-cut~\cite{ullman2023large, shapira2023clever}. Therefore, our work aims further to explore the capabilities of LLMs, specifically open-source models, contributing to a deeper understanding of their potential.

% Much prior work has focused on how LLMs can learn and be trained to become optimal decision-making agents compared to humans in multi-step goal-directed decision-making tasks from interactive environments~\cite{abdulhai2023lmrl,zhou2024archer}, and hence the question of  remains underexplored. 
% Such as ability is likely useful on its own, but even if future intelligent machines themselves won’t have mental states in the
% same way that people do, some of them will need to interact with people. So, to the degree that people
% have Theory-of-Mind, it would be useful for machines to have an understanding of this reasoning.

\section{Related Work}
\label{sec:related_work}
% This paper is closely related to the following three research domains:

\paragraph{LLMs for Agent Behavior Modeling.}
Generative agents use LLMs to drive their behavior, taking advantage of the extensive data on human behavior encoded in these models~\cite{brown2020language}. Research often relies on templates with few-shot prompts~\cite{gao2020making} or chain-of-thought prompts~\cite{wei2022chain} to effectively generate behavior based on the agent's environment. These templates have proven effective in the control and decision-making tasks. Recent work has shown that LLMs can produce human-like interactions in multi-player games involving natural language communication~\cite{park2023generative}. 
% This approach uses LLM prompts to select actions and facilitate agent communication, assessing the relevance and importance of information for decision-making.

Additionally, LLMs have been used to enhance agent modeling with reinforcement learning (RL) agents. Research has shown that integrating feedback into RL models through LLMs provides a learning experience similar to RL with human feedback, without requiring human judgments~\cite{wu2023spring,wu2023read,mcdonald2023exploring}. LLMs have also improved offline RL, reducing the need for computationally intensive online learning ~\cite{shi2023unleashing}.

We argue that LLMs offer an opportunity to leverage generative models for understanding and predicting human behavior. Unlike much existing work that models optimal AI agents, we focus on capturing human behavior.

\paragraph{LLMs in Theory of Mind Reasoning.}
A growing body of research has explored LLMs' Theory of Mind (ToM) capabilities by testing them with various ToM tasks~\cite{kosinski2023theory,strachan2024testing}. Results show that leading LLMs can solve 90\% of false-belief tasks, sometimes performing at or above human levels, indicating ToM-like abilities. However, ~\citet{ullman2023large,shapira2023clever} found that LLMs' performance deteriorates with slight modifications to task structure, highlighting mixed results in this area.

From a modeling perspective, ToM has been used to improve AI agent performance in different contexts. Recent studies have applied ToM with LLMs to enhance collaboration in multi-agent reinforcement learning~\cite{li2023theory,sun2024llm}. We distinguish our work by evaluating the ToM abilities of LLMs to understand human action strategies and biases across various decision-making complexities rather than focusing on learning to infer the intentions of other RL agents.

% In our work, we evaluate LLMs' ToM-like abilities by examining their alignment with real human action strategies and biases in varying decision-making complexities rather than focusing on learning to infer other RL agents' intentions.

\paragraph{Cognitive Modeling and Human Behavior.}
Cognitive architectures like ACT-R have demonstrated success in achieving human-level reasoning with limited training instances and in capturing cognitive biases in various decision-making tasks~\cite{anderson2014atomic, gonzalez2003instance, erev2010choice, thomson2015general}. \citet{lebiere2013functional} showed that the cognitive IBL model predicts whether a person will be risky or risk-averse based on previous trial feedback. Prior research has also indicated that IBL models align with human judgment in predicting RL agents' goals~\cite{nguyen2022theory} and often serve as a baseline for human behavior~\citep{malloy2023accounting}.

%Our work investigates how cognitive models, compared to LLMs, can predict and account for human action strategies based on past observations in decision-making processes. These processes involve balancing exploitative and exploratory actions while managing delayed feedback. By comparing the predictive performance of cognitive models with LLM-based models, we aim to establish a baseline for understanding their differences and potential synergies.

Building on research that behavioral traces predict performance~\citep{gadiraju2019crowd}, we explore how cognitive models, particularly the IBL model, compare to LLMs in predicting human action strategies in decision-making based on past observations. These processes involve balancing exploitative and exploratory actions with delayed feedback. By comparing the predictive performance of cognitive and LLM-based models, we aim to establish a baseline for understanding their differences and potential synergies.

\section{Preliminaries}
\label{sec:preliminary}
\paragraph{Task Scenario.} We studied goal-seeking task environments~\cite{rabinowitz2018machine,nguyen2023credit} that were set in $10 \times 10$ gridworld mazes containing obstacles and four terminal targets. Each target had a different value, with only one target having the highest value. The reward function over the four terminal objects was drawn randomly from a Dirichlet distribution with a concentration parameter of 0.01.
During each episode, the player navigated through the grid by making a series of decisions using the common action space (up, down, left, or right) to locate the target with the highest value. The player could consume the targets by moving on top of them. Episodes ended when a target was collected or when the time horizon was reached ($T_{\max} = 31$). The player received points for reaching the target but was also penalized for each movement (-0.01) and for walking into an obstacle (-0.05). 

\paragraph{Task Formulation.} The task is modeled as a partially observable Markov Decision Process (POMDP), represented by the tuple $\langle\mathcal{S}, \mathcal{A}, \mathcal{O}, \mathcal{T}, \mathcal{R}, \Omega, \gamma \rangle$. Here, $\mathcal{S}$ denotes the state space, with each square in the grid called a state $s \in \mathcal{S}$; $\mathcal{A}$ is the action space; $\mathcal{O}$ is the observation space; $\mathcal{T}: \mathcal{S} \times \mathcal{A} \to \mathcal{S}$ is the transition function; $\mathcal{R}: \mathcal{S} \times \mathcal{A} \to \mathbb{R}$ is the reward function; $\Omega: \mathcal{S} \to \mathcal{O}$ is the observation function; and $\gamma \in [0,1)$ is the discount factor controlling the player's emphasis on future rewards compared to immediate rewards.

At every step $t \in {0, ..., T_{\max}}$, a player is required to take an action $a \in \mathcal{A}$ after observing $o_t \in \mathcal{O}$. The player receives a reward $r_t \in \mathbb{R}$ after taking the action, as the environment transitions to a new state. Each player follows their policy $\pi_i$ (i.e., strategy) to decide how to act. y executing its policy $\pi_i$ in the gridworld $\mathcal{M}$ following episode $j$, the player $\mathcal{P}i$ generates a trajectory denoted by $\mathcal{T}{ij} = {(s_t, a_t)}_{t=0}^{T{\max}}$.

\section{Prediction Models}
\label{sec:study_design}
\begin{figure*}[!htbp]
\centering
\begin{subfigure}[b]{0.9\linewidth}
    \centering
    \includegraphics[width=0.9\linewidth]{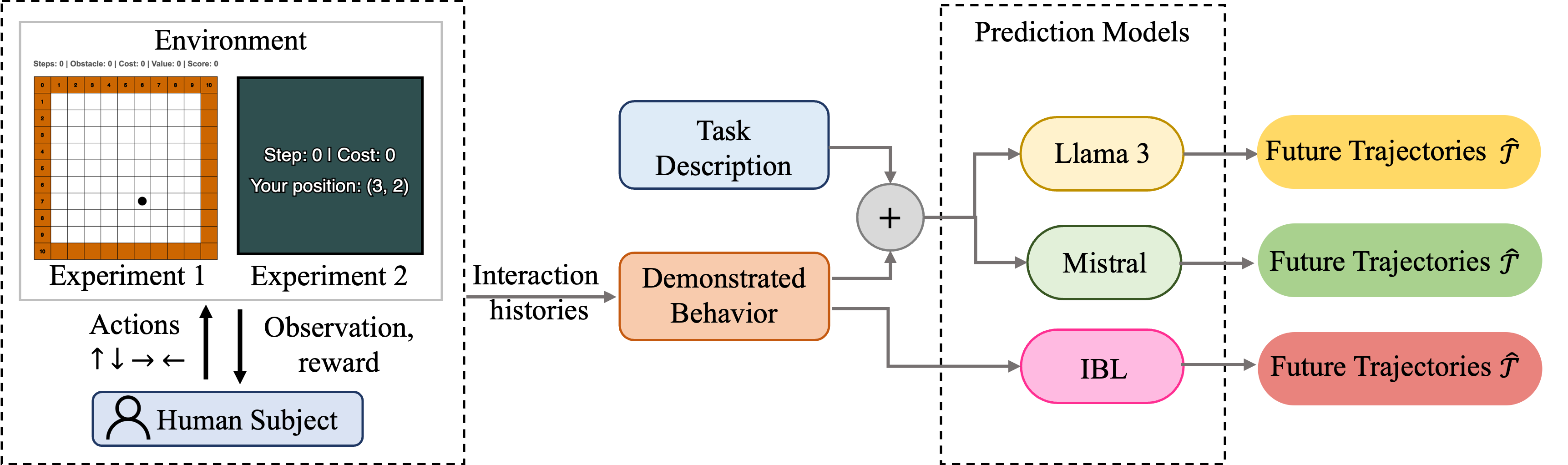}
    \caption{The IBL model used the demonstrated behaviors (past trajectories) to predict future behavior, specifically the next-episode trajectory. The LLMs used both the demonstrated behaviors and task descriptions to make such predictions.}
    \label{fig:design}
\end{subfigure}

\vspace{5mm} % Adjust the space between the subfigures as needed

\begin{subfigure}[b]{0.9\linewidth}
    \centering
    \includegraphics[width=\linewidth]{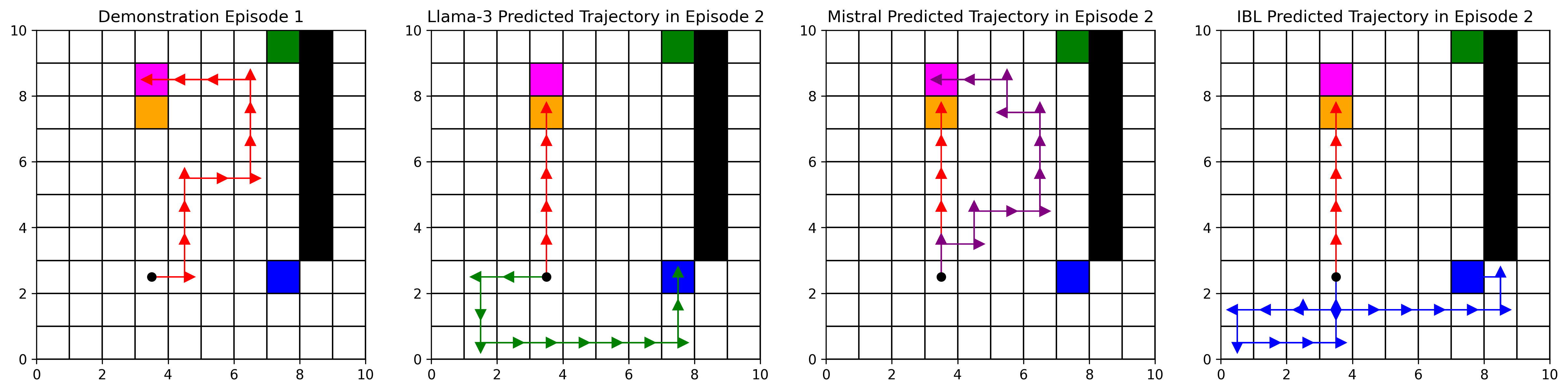}
    \caption{An example of the demonstrated trajectory in episode 1 (on the left-most plot), the models' predicted trajectories in the next episode, and the true human player trajectory in that episode (in red). Here, the orange target has the highest value.}
    \label{fig:demo}
\end{subfigure}
\caption{An overview of the experiment design.}
\label{fig:kl-experiment}
\end{figure*}

We describe a general approach that adapts LLMs for predicting human decisions in sequential goal-directed decision-making tasks. We compare our LLM-based prediction models with the cognitive IBL model. The overall framework of our approach is illustrated in Fig.~\ref{fig:design}.

% \begin{figure*}[!htbp]
% \centering
% \includegraphics[width=0.9\linewidth]{framework.png}
% \caption{An overview of the experiment design. The IBL model used the demonstrated behaviors (past trajectories of human players) to predict future behavior, specifically the next-episode trajectory. The LLMs used both the demonstrated behaviors and task descriptions to make such predictions.}
% \label{fig:experiment}
% \end{figure*}

\subsection{LLM-based Prediction Models}
We use LLMs to predict human behaviors in the described task, using both instruction-following and demonstration-based paradigms. Building on prior successes in using LLMs for control settings~\cite{wu2023read, wu2023spring, park2023generative}, we prompt the LLM to generate a trajectory that human players would take to succeed in the task. Unlike previous research~\cite{wu2023read, mcdonald2023exploring}, which focused on LLMs deciding the provision or value of a reward for optimal performance, we ask the LLM to predict the sequence of actions a human player would take. Specifically, we frame our query to predict the trajectory in the next episode, aiming to match the human strategy.

For each user, we construct a prompt consisting of two main parts: task instructions and sequential interaction histories. The task instructions (\texttt{$\langle$Task\_Instruction$\rangle$}) are detailed in the Appendix. The interaction histories include the starting position, the trajectory taken in previous episodes, and information about any consumed targets and their associated values. The prompt design is as follows:

% \begin{tcolorbox}[colback=black!5!white, colframe=black, width=\textwidth, boxrule=0.5mm, sharp corners, title=Task-wise Prompt Example]

\begin{tcolorbox}[colback=black!5!white, colframe=black, width=\columnwidth, boxrule=0.5mm, sharp corners, title=Prompt Design]

\textbf{Instruction:}
\texttt{\textit{$\langle$Task\_Instruction$\rangle$}.}

\textbf{Demonstration:}\\
\texttt{The (x,y)-coordinate of the starting position is \textit{$\langle$starting\_position$\rangle$}.}

\texttt{The trajectory of episode $j$: \textit{$\langle$trajectory\_j$\rangle$}.} \\
\texttt{The player collected goal \textit{$\langle$consumed\_goal\_j$\rangle$} with a score of \textit{$\langle$consumed\_value\_j$\rangle$}.} \\
\textbf{Template:} \\
\texttt{What is the trajectory the player would take in episode \textit{$\langle$j+1$\rangle$}? Please provide only the trajectory in the format of coordinate pairs [x,y]. Do not explain the reason or include any other words.}

\textbf{Output:} \texttt{\textit{$\langle$predicted\_trajectory$\rangle$}}.
\end{tcolorbox}

% As in the prompt, we construct the instruction using in-context learning (ICL), a prominent prompting approach for LLMs to solve various tasks~\cite{zhao2023survey}. This involves including demonstration examples in the prompt by augmenting the input interaction sequence. Specifically, we pair the prefix of the input interaction sequence with the corresponding successor as examples. For instance, ``\texttt{The trajectory of episode 1: $[(3, 2), (3, 3), (3, 4), (3, 5)]$. The player collected goal \textit{$blue$} with a score of \textit{$0.66$}}''. 
% This approach enables the LLMs to understand the instructions and output human behavior based on the sequential historical interactions provided.

In the prompt, we use in-context learning (ICL), a prominent prompting approach for LLMs to solve various tasks~\cite{zhao2023survey}. This involves incorporating demonstration examples by augmenting the input interaction sequence. Specifically, we pair the prefix of the input interaction sequence with its corresponding successor as examples. For instance, ``\texttt{The trajectory of episode 1: $[(3, 2), (3, 3), (3, 4), (3, 5)]$. The player collected goal \textit{$blue$} with a score of \textit{$0.66$}}''. 
To predict the next actions, we ask, ``\texttt{What is the trajectory the player would take in episode \textit{$2$?}}''. This approach enables the LLMs to understand the instructions and output human behavior based on their sequential historical interactions provided.

\paragraph{POMDP Formulation.} The LLM-based prediction model can be formalized as follows: For each player $\mathcal{P}i$, we aggregate their historical trajectories $\mathcal{T} = \bigcup_{j} \mathcal{T}{ij} = {(s_t, a_t)}_{t=0}^{T_{\max}}$, generated after executing a sequence of actions at each time step $t$ across $j$ episodes.

The context $\mathcal{C}$ is defined by the combination of task instructions and historical interactions $\mathcal{T}$, paired with the consumed targets $\mathcal{G} = \{g_1, \ldots, g_j\}$ corresponding to the trajectory of each episode $j$, and the values $\mathcal{V} = \{v_1, \ldots, v_j\}$ associated with obtaining these targets. The LLM model $M$ uses this context $\mathcal{C}$ as input to predict the trajectory for the next episode $\mathcal{T}_{i(j+1)}$. Essentially, given the context $\mathcal{C}$ spanning from the first to the $j$-th episode, the LLM-based model predicts the trajectory for episode $j+1$. The details of the algorithm are provided in Algorithm~\ref{alg:algorithm}.

\begin{algorithm}
\caption{Trajectory Prediction Using LLM}
\begin{algorithmic}[1]
\STATE Initialize result storage as a dictionary $\texttt{result} = \{\}$
\FOR{each player $\mathcal{P}_i$}
    \STATE Initialize the context $\mathcal{C}$ with task instructions and starting position
    \FOR{each episode $j$}
        \STATE Aggregate historical trajectories $\mathcal{T}_{ij}$ and corresponding consumed targets $\mathcal{G}_j$ and values $\mathcal{V}_j$
        \STATE Update the context $\mathcal{C}$ with $\mathcal{T}_{ij}$, $\mathcal{G}_j$, and $\mathcal{V}_j$
        \STATE Query the LLM model $M$ with context $\mathcal{C}$ to predict the next episode's trajectory $\mathcal{T}_{i(j+1)}$
        \STATE Store the predicted trajectory $\mathcal{T}_{i(j+1)}$ in $\texttt{result}$
    \ENDFOR
\ENDFOR
\end{algorithmic}
\label{alg:algorithm}
\end{algorithm}

We note that the output from LLMs may still contain natural language text. We address this by employing text processing methods to parse and ground the generated results in the specified environment. Additionally, we have occasionally observed instances where LLMs produce coordinates that are invalid within the given environment scope. In such cases, we reprocess these illegal outputs to ensure compliance with environmental constraints.

% \paragraph{Parsing the Output of LLMs.} It is important to note that the output from LLMs may still contain natural language text. We address this by employing text processing methods to parse and ground the generated results in the specified environment. Additionally, we have occasionally observed instances where LLMs produce coordinates that are invalid within the given environment scope. In such cases, we reprocess these illegal outputs to ensure compliance with environmental constraints.

\subsection{Instance-Based Learning (IBL) for Prediction}
The IBL model used for comparing and predicting human 
behavior is based on Instance-Based Learning Theory (IBLT)~\cite{GONZALEZ03} for dynamic decision-making, which is connected to the ACT-R cognitive architecture through the activation function, which is used to predict the estimated utility of performing an action in a state based on the utility outcomes of similar past experiences held in declarative memory \cite{thomson2015general}. 

In IBLT, declarative memory consists of instances $k = (o,a,x)$ represented by the observation that describes the state of the environment $o$, the action performed by the agent $a$, and the utility outcome of that action $x$. This instance structure can be related to the POMDP environment formulation by taking the state $s$ to be the agent observation $o$, and the utility outcome $x$ to be the observed reward $r$. 

Agent actions are determined by maximizing the value $V_{k,t}$ of an available action $a$ in an instance $k$ performed at time-step $t$, calculated using the ``blending'' function \cite{GONZALEZ03}:
\begin{equation}
    V_{k,t} = \sum_{i=1}^{n_{k,t}}p_{i,k,t} x_{i,k,t}
\label{eq:Blending}
\end{equation}
where $n_{k,t}$ are the previously generated instances held in procedural memory, $x_{i,k,t}$ are the outcomes of those instances, and $p_{i,k,t}$ is the probability of retrieving an instance in memory, calculated by Equation~\ref{eqn:prob_retrieval}.
\begin{equation} \label{eqn:prob_retrieval}
    p_{i,k,t} = \frac{\exp(\Lambda_{i,k,t}/\tau)}{\sum_{j=1}^{n_{k,t}} \exp(\Lambda_{j,k,t}/\tau)}
\end{equation}
Further, $\Lambda_{i,k,t}$ is given by Equation~\ref{eqn:lambda}.
\begin{equation} \label{eqn:lambda}
\Lambda_{i,k,t} = \ln \left( \sum_{t' \in T_{i,k,t}} (t - t')^{-d} \right) + \sigma \ln \frac{1 - \xi_{i,k,t}}{\xi_{i,k,t}},
\end{equation}
where $d$ and $\sigma$ are decay and noise parameters, and $T_{i,k,t} \subset \{0,...,t-1\}$ is the set of previous timesteps where instance $k$ was stored in memory. The $\xi_{i,k,t}$ term is used to capture noise in the individual differences in memory recall. Because of the relationship between noise $\sigma$ and temperature $\tau$ in IBLT, the temperature parameter $\tau$ is typically set to $\sigma \sqrt{2}$. In our experiments, we use all default parameters of $d=0.25$ and $\sigma=0.5$. We also set the default utility to 1.0 to encourage exploration through an optimistic prior.%~\cite{sutton2018reinforcement}. 
% The default utility is used to predict the utility of an instance when there are no similar instances in memory to estimate the expected utility. 

A key aspect of applying IBLT to decision-making is determining the utility of actions. Prior research on temporal credit assignment in IBL models has shown that models assigning equal credit to all decisions closely match human performance~\cite{nguyen2023credit,nguyen2024temporal}, which we consequently have chosen to adopt this approach. Formally, if a target is reached at step $T$, the target's value $R_T$ is assigned to each instance in the trajectory $\mathcal{T}={(s_t, a_t)}_{t=0}^{T}$, i.e., $x_t = R_{T}$ for all $(s_t, a_t)$. The step-level costs are assigned to each instance if no target is reached.

The IBL prediction model, functioning as an observer, learns by observing past decisions made by human agents. This past experience is incorporated into the model's memory through pre-populated instances, a mechanism that demonstrates how the IBL model can dynamically represent the development of ToM by observing actions of other learning agents in a gridworld task~\cite{nguyen2022theory}. Specifically, for each player $\mathcal{P}i$, the trajectory $\mathcal{T}_{ij}$ produced by the player, following its policy $\pi$ in the gridworld $\mathcal{M}$ after episode $j$, is stored in the model memory.

\section{Methods}
\label{sec:study_design}

% In Experiment 1, human participants saw their current position in the grid (a black dot). After each move, the new location's content (empty cell, obstacle, or target) was revealed. In Experiment 2, participants had limited information, seeing only one cell at a time, with the grid's shape and size concealed. 

Our research aims to determine if LLMs and the cognitive IBL model can accurately predict human strategic planning in uncertain decision-making environments, formulated as POMDPs, given past interaction histories. These environments require balancing potential high rewards against the risks or losses associated with high-value objects.

We utilized data from two human-subject experiments using interactive browser-based gridworld applications, which incorporate two levels of decision complexity. Moreover, we explored how the models predict human strategies under different levels of environment presentation: full grid information in Experiment 1 and restricted grid information in Experiment 2. Our primary research questions are:
\begin{itemize}
    \item \textbf{RQ1}: How accurately can LLMs and the cognitive IBL model predict human action in uncertain decision-making environments based on past interaction histories?
    \item \textbf{RQ2}: To what extent do LLMs and the cognitive IBL model capture human decision biases, such as loss aversion, across different levels of decision complexity?
    \item \textbf{RQ3}: How do different levels of environment presentation (full grid information vs. restricted grid information) affect the accuracy of LLMs and the cognitive IBL model in predicting human decision strategies?

\end{itemize}
\subsection{Experimental Design and Procedure}
The two experiments used the same gridworlds, but the information provided to participants varied. Human subjects were presented with gridworlds randomly chosen from a set of 100 grids, with the selection based on the decision complexity level assigned to each participant's condition.

Participants were recruited from Amazon Mechanical Turk and provided informed consent before completing each session. After receiving instructions, participants completed 40 episodes in the same gridworld environment, with each session lasting 15-30 minutes. They received a base payment of \$1.50 and could earn up to \$3.00 in bonuses based on their accumulated scores. The studies, approved by our institution's IRB, employed a between-subjects design and were preregistered with the Open Science Framework for Experiment 1\footnote{Experiment 1: \url{https://osf.io/2ycm6}} and Experiment 2\footnote{Experiment 2: \url{https://osf.io/hxfyq}}.

\begin{figure}[!htbp]
\centering
 \begin{subfigure}[b]{0.48\linewidth}
        \includegraphics[width=\linewidth]{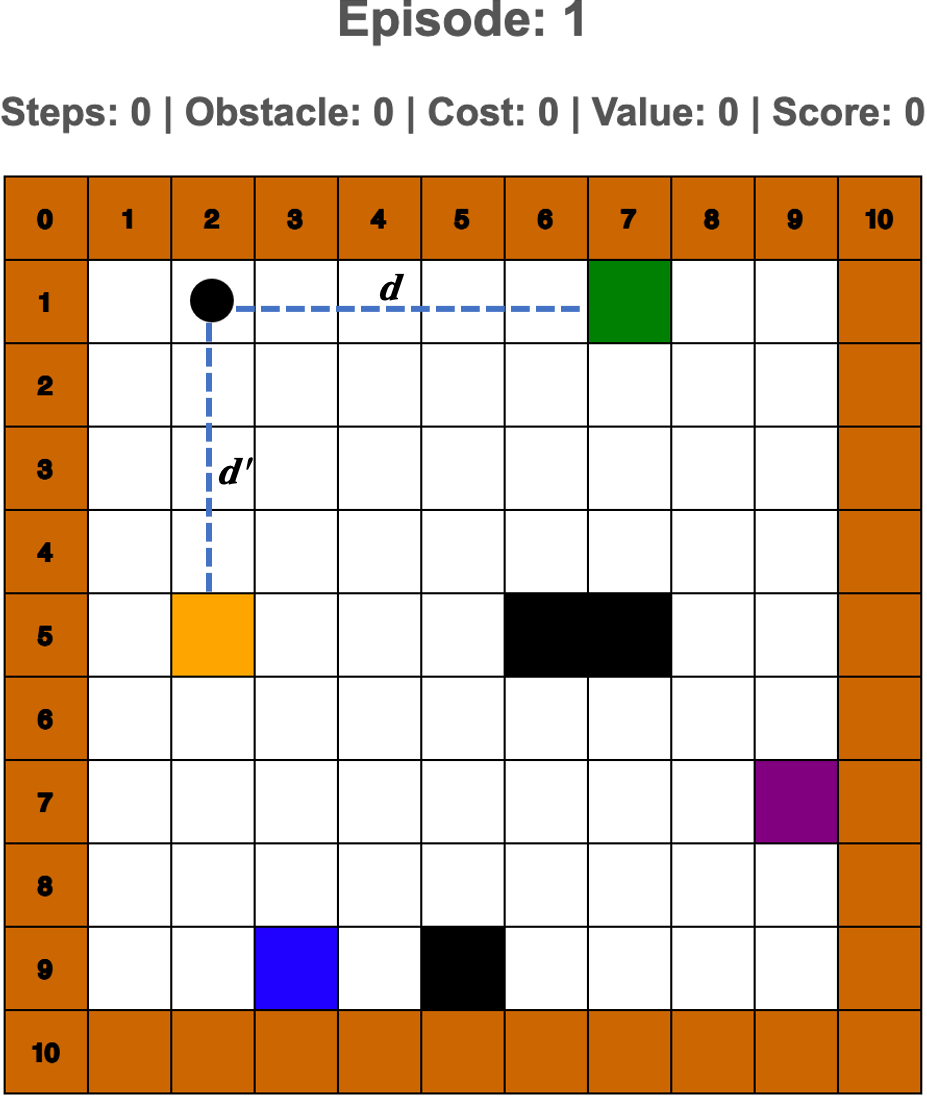}
\caption{\textbf{Simple grid}}
\label{fig:simple}
    \end{subfigure}\hspace{1mm} %or \hspace{0.3\textwidth}
\begin{subfigure}[b]{0.48\linewidth}
        \includegraphics[width=\linewidth]{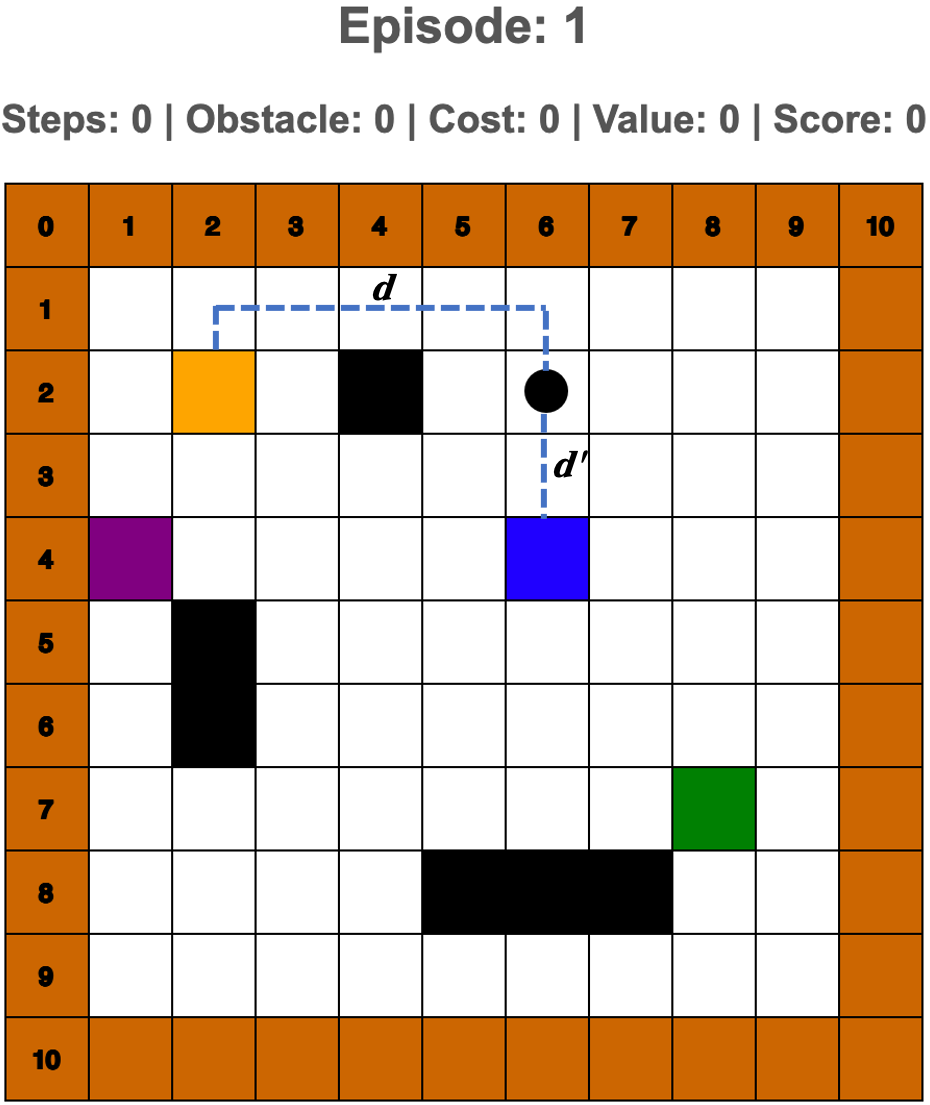}
\caption{\textbf{Complex grid}}
\label{fig:complex}
    \end{subfigure} 
\caption{Example grids for simple and complex conditions:(a) ``green'' is the highest value target with ``orange'' as the distractor ($\Delta_d = 1$); (b) ``orange'' is the target, ``blue'' is the distractor ($\Delta_d = 4$).}
\label{fig:conditions}
\end{figure}

\paragraph{Decision Complexity.}
The experiments manipulate the level of decision complexity defined by the trade-offs between the highest value target and the nearest distractor relative to the agent's initial spawn location in the gridworld~\cite{nguyen2020effects}. This complexity is quantified by $\Delta_d = d - d'$, where $d$ is the distance to the highest value target, and $d'$ is the distance to the nearest distractor. Higher $\Delta_d$ values represent greater complexity, posing a strategic dilemma to agents: pursue a distant high-reward target or opt for a closer, less valuable one. Fig.~\ref{fig:conditions} illustrates simple and complex decision scenarios.

% The exact instructions for both experiments are available in the Supplementary Materials.

\paragraph{Experiment 1: Full Grid Information.}

Participants viewed the full grid interface (Fig.\ref{fig:design}), with their current position indicated by a black dot. After each move, the new location content (empty cell, obstacle, or target) was revealed.

\textit{Participants.} A total of 206 participants: 102 in the Simple condition (age: 36.5 $\pm$ 10.3; 34 female) and 104 in the Complex condition (age: 37.9 $\pm$ 10.8; 37 female).

\paragraph{Experiment 2: Restricted Grid Information.}
Participants received limited information, viewing only one cell at a time (Fig.~\ref{fig:design}). They were informed of their current ($x$, $y$) position, the steps taken, and the immediate cost or reward of the previous step. All other information about the shape and size of the grid were concealed.

\textit{Participants.} A total of 194 participants: 99 in the Simple condition (age: 37.7 $\pm$ 11.8; 40 female) and 95 in the Complex condition (age: 38.2 $\pm$ 11.3; 30 female).

\paragraph{Model Implementation.} We implemented the cognitive IBL model using the SpeedyIBL library~\citep{nguyen2023speedyibl} with default parameters. The LLM-based models were accessed through the Ollama REST API\footnote{\url{https://github.com/ollama/ollama}}.

\subsection{Objective Measures}
\begin{itemize}
    % \item \textbf{Regret}: The difference between the value of the highest value target and the value of the target actually reached by the player or predicted by the model.

    \item \textbf{Trajectory Divergence}: We used Kullback-Leibler (KL) divergence to measure the difference between the trajectory distribution of human subjects and that predicted by the model. Each trajectory is converted into a probability distribution over the grid cells, normalized to sum to 1. The KL divergence from $Q$ (predicted trajectory) to $P$ (human trajectory) is defined as: 
    \begin{equation}
    D_{KL}(P \| Q) = \sum_{i} P(i) \log \frac{P(i)}{Q(i)}
    \end{equation}
    
    where \( i \) indexes each possible state in the trajectory grid. Low KL divergence indicates that the model closely matches human behavior. %with a value of 0 meaning identical distributions.

    \item \textbf{Prediction Accuracy}: This measures  the percentage of episodes where the predicted target, derived from the last coordinate of the predicted trajectory, matches the target consumed by human players.

    \item \textbf{Exploration Entropy Difference}: This metric measures the difference in entropy of the distribution over how often each target is explored by humans (human goal entropy) and by the model (predicted goal entropy) in the first 10 episodes. The entropy difference is determined by subtracting the human goal entropy from the predicted goal entropy. Lower entropy difference suggests that the model's exploration behavior closely aligns with human behavior, indicating similar patterns in exploring targets. %This approach provides insight into how well the models replicate human exploratory tendencies and risk-taking behavior.

    % \item \textbf{Exploration Entropy}: This metric measures the entropy of the distribution over how often each target is explored in the first half of episodes. A perfectly greedy policy, which always selects the same target, has an entropy of 0, while a perfectly uniform policy, which explores all 4 targets equally, has an entropy of 2. Higher entropy indicates more varied exploration, suggesting a willingness to take risks.

\end{itemize}

\section{Analysis}
\label{sec:results}
%We compared two LLM-based prediction models, Mistral 7B~\cite{jiang2023mistral} and Llama-3 70B~\cite{meta_llama_3}, and the cognitive IBL model across Experiments 1 and 2. Each experiment included two levels of decision complexity: simple and complex, resulting in a $2 \times 2$ study design.

%We compared the LLM-based models Llama-3 70B~\cite{meta_llama_3} and Mistral 7B~\cite{jiang2023mistral} with the cognitive IBL model across a $2 \times 2$ study design in two experiments, each with simple and complex decision complexities.

We compared the LLM-based models Llama-3 70B~\cite{meta_llama_3} and Mistral 7B~\cite{jiang2023mistral} with the cognitive IBL model across a $2 \times 2$ study design in two experiments, each with simple and complex decision complexities.

\paragraph{Trajectory Divergence.}
Table~\ref{tab:avg-kl} shows that in both experiments, the Mistral model with 7B consistently achieved the lowest KL divergence in simple and complex conditions, indicating better alignment with human trajectories compared to Llama-3 with 70B and the cognitive IBL model. Comparing simple and complex decision settings, predicting human strategies is more challenging in complex environments, as evidenced by the increased KL divergence. 

When comparing Experiments 1 and 2, we observe that both Llama-3 and Mistral show better alignment with human trajectories under restricted information, as indicated by decreased KL divergence in Experiment 2, while this is not the case for the IBL model. This improved alignment for the LLMs can be attributed to their pre-training on vast datasets and the incorporation of instructions, which enhance their contextual understanding. When human participants receive restricted grid information, as in Experiment 2, they may adopt more predictable strategies that LLM models can more easily capture. By contrast, the IBL model shows poorer alignment with human trajectories under restricted information as it struggles to distinguish the differences that human players encountered in the two different conditions.

% while this is not the case for the IBL model. This suggests that the LLM models are more effective at capturing human strategies when the human subjects are presented with limited grid information. 
 % This suggests that the IBL model relies more on full environmental context to make accurate predictions. 

% By contrast, in comparing the IBL model with Llama-3 and Mistral models, we observe distinct trends across the two experiments. Both Llama-3 and Mistral show a decrease in KL divergence in Experiment 2, indicating better alignment with human trajectories under restricted information. This suggests that these LLMs adapt well to limited data.
% The results of the experiments highlight distinct differences in the performance of various models under simple and complex conditions across two experiments. 

% In Experiment 1, Mistral consistently achieved the lowest KL divergence in both simple and complex conditions, indicating a better alignment with human trajectories compared to Llama-3 and IBL. Specifically, Mistral recorded a KL divergence of 4.895 in the simple condition and 6.524 in the complex condition, outperforming Llama-3 and IBL, which showed higher values in both scenarios. Similarly, in Experiment 2, Mistral continued to demonstrate superior performance with a KL divergence of 4.641 in the simple condition and 5.910 in the complex condition, again outperforming Llama-3 and IBL. 
% These results suggest that Mistral is more effective at predicting human strategies, particularly in more challenging environments. 

\begin{table}[ht]
\centering
\caption{Average KL Divergence for each model under different conditions in Experiment 1 and Experiment 2.}
\begin{tabular}{lllc}
\hline
\textbf{Experiment} & \textbf{Condition} & \textbf{Model} & \textbf{Mean KL $\pm$ SE} \\
\hline
\multirow{6}{*}{Experiment 1} & \multirow{3}{*}{Simple}   & Llama-3 & 6.189 $\pm$ 0.128 \\
                              &                            & \textbf{Mistral} & \textbf{4.895 $\pm$ 0.117} \\
                              &                            & IBL     & 5.244 $\pm$ 0.122 \\
                              & \multirow{3}{*}{Complex}  & Llama-3 & 8.385 $\pm$ 0.126 \\
                              &                            & \textbf{Mistral} & \textbf{6.524 $\pm$ 0.118} \\
                              &                            & IBL     & 7.216 $\pm$ 0.119 \\
\hline
\multirow{6}{*}{Experiment 2} & \multirow{3}{*}{Simple}   & Llama-3 & 5.901   $\pm$ 0.127 \\
                              &                            & \textbf{Mistral} & \textbf{4.641   $\pm$ 0.114} \\
                              &                            & IBL     & 6.061 $\pm$ 0.128 \\
                              & \multirow{3}{*}{Complex}  & Llama-3 & 7.416   $\pm$ 0.129 \\
                              &                            & \textbf{Mistral} & \textbf{5.910   $\pm$ 0.120} \\
                              &                            & IBL     & 8.853   $\pm$ 0.123 \\
\hline
\end{tabular}
\label{tab:avg-kl}
\end{table}

\begin{figure}[!htbp]
\centering
 \begin{subfigure}[b]{1\linewidth}
        \includegraphics[width=\linewidth]{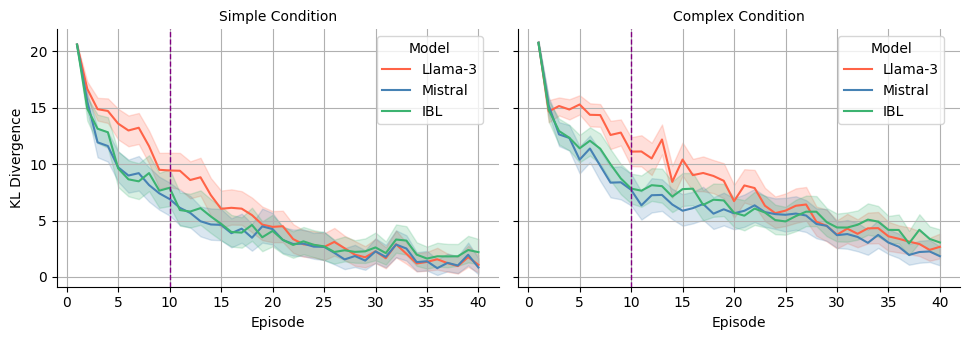}
\caption{Experiment 1 - Full Grid Information}
\label{fig:exp1-kl}
    \end{subfigure}\hspace{1mm} %or \hspace{0.3\textwidth}
\begin{subfigure}[b]{1\linewidth}
        \includegraphics[width=\linewidth]{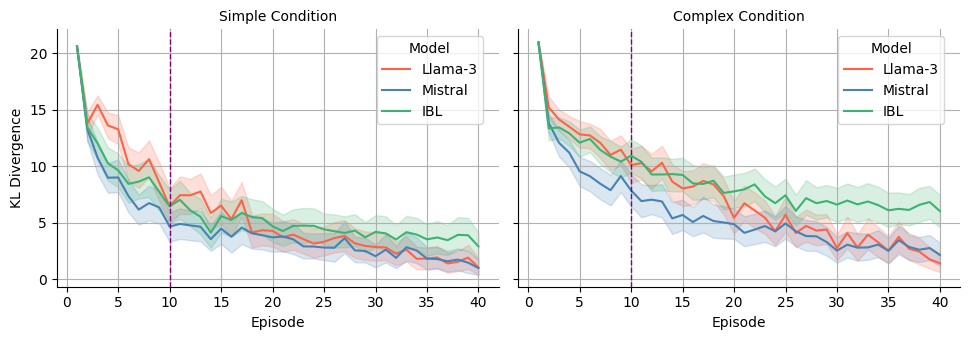}
\caption{Experiment 2 - Restricted Grid Information}
\label{fig:exp2-kl}
    \end{subfigure} 
\caption{Average KL divergence per episode for all models in both conditions of Experiments 1 and 2, with shaded areas indicating standard error at 95\% confidence intervals. Lower KL divergence suggests better alignment.}
\label{fig:kl-episode}
\end{figure}

Fig.~\ref{fig:kl-episode} shows the average KL divergence per episode. As expected, in both experiments and conditions, all three models show a decreasing KL divergence trend, indicating better alignment with human trajectories as episodes and demonstrated samples increase. Notably, in the simple condition, the IBL model performs comparably to Mistral and even outperforms Llama-3 in the first 10 episodes, suggesting effective capturing of human actions with few-shot examples.

% In experiment 1, Llama-3 consistently exhibits the highest KL divergence across all conditions, indicating the poorest alignment with human trajectories. Llama-3 shows the highest KL divergence overall, making it the worst performer among the three models. 

\paragraph{Prediction Accuracy.}
The results from Table~\ref{tab:avg-acc} indicate that the cognitive IBL model is the most accurate in predicting human target consumption in simple conditions in Experiment 1, achieving the highest prediction accuracy (0.688 $\pm$ 0.007) compared to Llama-3 and Mistral.

In both experiments, all models exhibit lower prediction accuracy in complex settings compared to simple settings. In Experiment 2, while Llama-3 and Mistral maintain similar accuracies to those in Experiment 1, the IBL model shows a decrease in prediction accuracy, especially in complex conditions. This suggests that the IBL model may struggle with capturing the high variability of human decisions when participants are provided with limited information.

\begin{table}[ht]
\centering
\caption{Average Prediction Accuracy for each model under different conditions in Experiment 1 and Experiment 2.}
\begin{tabular}{lllc}
\hline
\textbf{Experiment} & \textbf{Condition} & \textbf{Model} & \textbf{Accuracy $\pm$ SE} \\
\hline
\multirow{6}{*}{Experiment 1} & \multirow{3}{*}{Simple}   & Llama-3 & 0.651 $\pm$ 0.007 \\
                              &                            & Mistral & 0.658 $\pm$ 0.007 \\
                              &                            & \textbf{IBL}     & \textbf{0.688 $\pm$ 0.007} \\
                              & \multirow{3}{*}{Complex}  & Llama-3 & 0.523 $\pm$ 0.008 \\
                              &                            & \textbf{Mistral} & \textbf{0.594 $\pm$ 0.008} \\
                              &                            & IBL     & 0.560 $\pm$ 0.008 \\
\hline
\multirow{6}{*}{Experiment 2} & \multirow{3}{*}{Simple}   & Llama-3 & 0.652 $\pm$ 0.008 \\
                              &                            & \textbf{Mistral} &\textbf{ 0.655 $\pm$ 0.008} \\
                              &                            & IBL     & 0.608 $\pm$ 0.008 \\
                              & \multirow{3}{*}{Complex}  & Llama-3 & 0.542 $\pm$ 0.008 \\
                              &                            & \textbf{Mistral} & \textbf{0.602 $\pm$ 0.008} \\
                              &                            & IBL     & 0.382 $\pm$ 0.008 \\
\hline
\end{tabular}
\label{tab:avg-acc}
\end{table}

\begin{figure}[!htbp]
\centering
 \begin{subfigure}[b]{1\linewidth}
        \includegraphics[width=\linewidth]{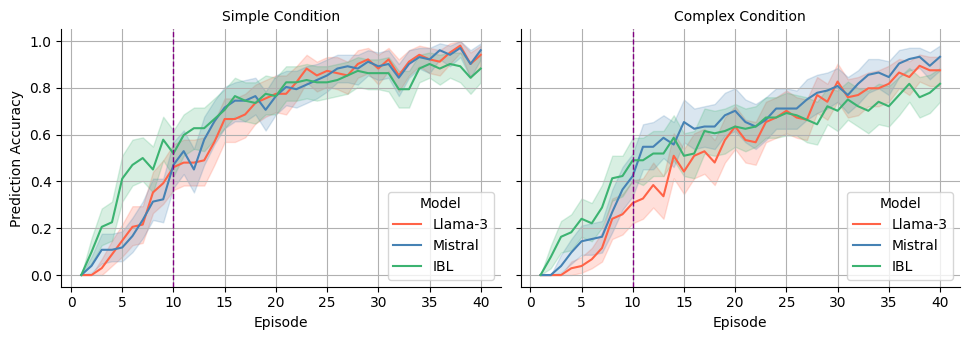}
\caption{Experiment 1 - Full Grid Information}
\label{fig:exp1-kl}
    \end{subfigure}\hspace{1mm} %or \hspace{0.3\textwidth}
\begin{subfigure}[b]{1\linewidth}
        \includegraphics[width=\linewidth]{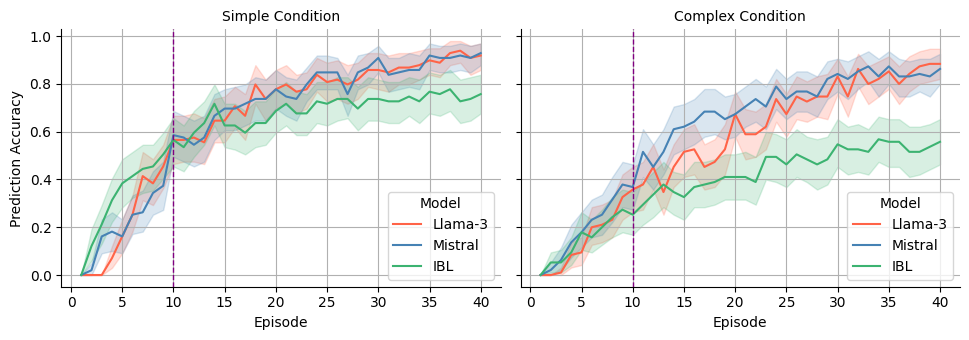}
\caption{Experiment 2 - Restricted Grid Information}
\label{fig:exp2-kl}
    \end{subfigure} 
\caption{Average prediction accuracy per episode for all models in both conditions of Experiments 1 and 2. Shaded areas indicate the standard error. Higher prediction accuracy suggests better alignment with human target consumption.}
\label{fig:acc-episode}
\end{figure}

It is noteworthy that in the first 10 episodes, the IBL model shows high prediction accuracy, surpassing both Llama-3 and Mistral in both simple and complex conditions, as shown in Fig.~\ref{fig:acc-episode}. This suggests that the IBL model effectively captures the initial exploration strategies of human players, learning quickly and adapting well with few samples. However, as more episodes are added, the LLMs pick up and quickly improve their prediction accuracy, eventually matching or exceeding the performance of the IBL model. 

This finding indicates that while the IBL model aligns closely with human decisions during initial exploration, LLMs can leverage larger amounts of data to refine their predictions and better capture human actions over time.

\begin{figure}[!htbp]
\centering
 \begin{subfigure}[b]{1\linewidth}
        \includegraphics[width=\linewidth]{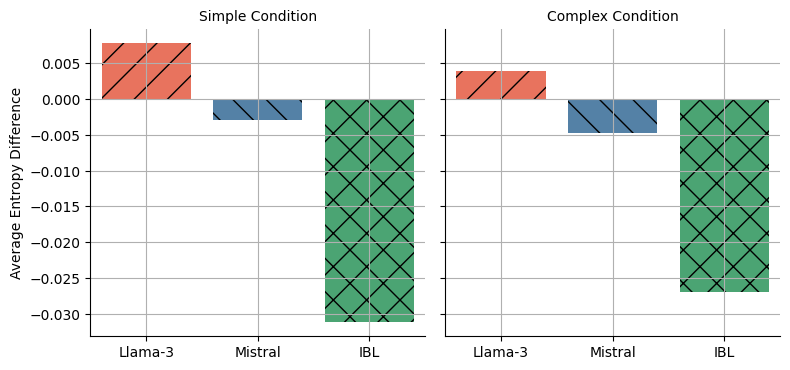}
\caption{Experiment 1 - Full Grid Information}
\label{fig:exp1-kl}
    \end{subfigure}\hspace{1mm} %or \hspace{0.3\textwidth}
\begin{subfigure}[b]{1\linewidth}
        \includegraphics[width=\linewidth]{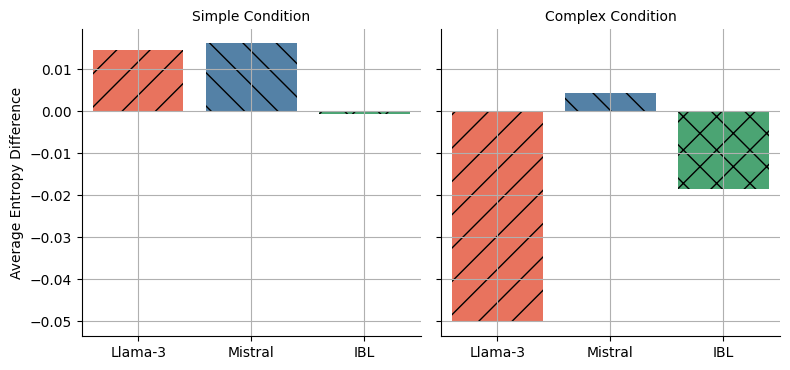}
\caption{Experiment 2 - Restricted Grid Information}
\label{fig:exp2-kl}
    \end{subfigure} 
\caption{Average entropy difference for all models in both conditions of Experiment 1 and 2. Positive values indicate more exploration than humans; negative values indicate less and near-zero values show alignment with human behavior.}
\label{fig:entropy-diff}
\end{figure}

\paragraph{Exploration Entropy Difference.}
Fig.~\ref{fig:entropy-diff} shows the differences in early exploratory behavior between each model and humans within the first 10 episodes. In Experiment 1, IBL shows a negative entropy difference in both simple and complex conditions, which suggests that it explores less than humans. On the other hand, Llama-3 and Mistral exhibit entropy differences close to zero or slightly positive, with Mistral aligning most closely with human exploration behavior.

% In Experiment 2, IBL again shows negative or near-zero entropy differences, especially in simple conditions, indicating conservative exploration. Interestingly, this exploration characteristic is in close alignment with human behavior, which suggests that in Experiment 2, with restricted information, people tend to adopt risk-averse strategies, well captured by the IBL model. 

In Experiment 2, IBL again shows negative or near-zero entropy differences, particularly in simple conditions, indicating conservative exploration. This aligns with human behavior, suggesting that with restricted information, people adopt risk-averse strategies well captured by the IBL model.

% It is worth noting that Mistral and IBL show similar trends in exploration behavior under simple and complex conditions. However, Llama-3 displays a slightly negative difference in the complex condition and a slightly positive difference in the simple condition. This suggests that Llama-3 is sensitive to and attempting to adapt to the observed behavior.

The results indicate that the IBL model, with default parameters, naturally exhibits a conservative exploration strategy, aligning with prior findings regarding its risk-aversion tendency compared to winner-take-all instances, which is the optimal strategy overall \citep{lebiere2013functional}. This characteristic effectively captures human exploration behavior when information is restricted. In such settings, people are limited in their exploration and often settle on the closest sub-optimal target rather than seeking the highest-value target, reflecting a human bias towards risk aversion. In contrast, Mistral consistently demonstrates exploratory patterns that are more aligned with those of humans.

\section{Discussions}
\label{sec:discussion}

In this paper, we investigate whether large language models (LLMs) can predict human action strategies and capture human biases, such as loss aversion, in decision-making scenarios involving cost-reward tension. We used two state-of-the-art open-source LLMs (Llama-3 70B and Mistral-7B) and compared them to a well-known cognitive instance-based learning (IBL) model. We tested these models in two experimental studies where human participants engaged in multi-step decision-making tasks in interactive environments with varying levels of information presentation. 

Our results show that Mistral-7B outperforms both Llama-3 70B and the cognitive IBL model in predicting human strategies. We also found that predicting human behavior becomes more challenging in complex decision environments with high tension between costs and rewards. Moreover, the cognitive IBL model effectively captures initial human exploratory behavior with minimal demonstration samples. However, as more samples are provided, LLMs quickly improve their prediction accuracy by leveraging their extensive pre-training on vast datasets and instructions that enhance their contextual understanding. Importantly, our findings indicate that the IBL model, with its inherent risk-aversion tendency, closely aligns with human exploration strategies under limited information conditions. It effectively captures the human tendency toward risk-averse, ``satisficing'' behavior~\citep{simon1956rational}, where people often choose an option that is satisfactory rather than optimal.

\subsection{Implications for Trust and Synergy in AI-assisted Systems with LLMs and Cognitive Models}
%One of our key findings is that a lightweight LLM, Mistral-7B, performs better than the widely recognized Llama-3 70B and the cognitive IBL model in accurately predicting and capturing human action strategies across various decision-making settings. This result highlights the potential of leveraging these open-source, lightweight LLMs to develop more reliable and trustworthy AI systems in decision-making contexts. While much of the extant research on LLM-powered human interaction focuses on investigating black-box LLMs, such as ChatGPT, which only provides an accessible API, our study emphasizes the potential capabilities of open-source pre-trained LLMs. These models enable fine-tuning and provide full access to network architecture and pre-trained weights, facilitating cognitive-plausible understanding and integration~\cite{binz2023turning, malloy2024applying}. Furthermore, prior research indicates that humans initially under-trust AI and tend to over-trust it with more experience, underscoring the need for explicit calibration to the AI's competencies~\cite{rechkemmer2022confidence, buccinca2021trust}. By leveraging these open-source models, we can create trustworthy LLM-powered systems that are more accurate, explainable, and aligned with human expectations, thereby enhancing their acceptance and effectiveness in real-world applications.

One of our findings is that the lightweight LLM Mistral-7B outperforms the widely recognized Llama-3 70B and the cognitive IBL model in predicting human action strategies across various decision-making settings. This highlights the potential of using open-source, lightweight LLMs to develop reliable AI systems. While much research on LLM-powered human interaction focuses on black-box models like ChatGPT, our study emphasizes the capabilities of open-source pre-trained LLMs, which allow fine-tuning and provide access to network architecture and weights, facilitating cognitive-plausible understanding and integration~\cite{binz2023turning, malloy2024applying}. Prior research shows humans initially under-trust AI and then over-trust it with more experience, highlighting the need for explicit calibration of AI competencies~\cite{rechkemmer2022confidence, buccinca2021trust}. Leveraging open-source models can create trustworthy, accurate, explainable, and aligned LLM-powered systems, enhancing their acceptance and effectiveness in real-world applications.

Our experiments with POMDPs, focused on simplicity and ease of control, have potential to generalize to richer domains where decisions involve uncertainty and incomplete information. In the context of AI-assisted human decision-making, our findings highlight the utility of the cognitive IBL model in capturing initial human exploratory behavior and the tendency towards loss aversion in high cost-reward tension scenarios without requiring large amounts of training data. Conversely, LLMs can quickly learn and predict human strategies as more data becomes available, indicating a synergy between these models. For instance, LLMs can support cognitive models by synthesizing large amounts of information and serving as knowledge repositories to construct representations of the environment~\cite{wu2023read,binz2023turning}.
Cognitive IBL models, which can predict human strategies with few-shot learning, 
have proven beneficial when integrated with multi-agent deep reinforcement learning techniques to enhance coordination in multi-agent systems with stochastic rewards~\cite{nguyen2023learning}. Thus, cognitive models can help LLMs adapt early on and further personalize their responses to human users~\cite{malloy2024applying,thomson2023integrating}. This synergy would benefit effective human-AI teaming, where AI evolves alongside human learning and adaptation to support human decision-making.

%Cognitive models, which can predict human performance and preferences with few-shot learning, can help LLMs adapt early on and further personalize their responses to human users~\cite{malloy2024applying,thomson2023integrating}. This synergy would benefit for effective human-AI teaming, where AI evolves alongside human learning and adaptation to support human decision-making.

\subsection{Limitations and Future Work}
Our experiments were simple and aimed to shed light on the capabilities of open-source LLMs and cognitive architectures in predicting human behavior. There is considerable room for further investigation. 
First, our study focused on purely open-source LLMs, so caution should be exercised when extrapolating our findings to closed-source commercial services like ChatGPT, which may exhibit different performance levels. Second, we used the vanilla versions of these models without fine-tuning them to human data, particularly in cognitive IBL models, where an equal credit assignment mechanism leads to conservative exploratory strategies. Future research could enhance these models to better match human behavior and uncover new behavioral structures. Finally, while our study highlights the strengths of different models, it does not fully explore the integration of LLMs and cognitive models in a cohesive framework, which we consider as future work. Future research would also benefit from investigating the potential of these models in interactive systems to predict human decisions in real time.

%First, our study focused on purely open-source LLMs, so caution should be exercised when extrapolating our findings to closed-source commercial services like ChatGPT, which may exhibit different performance levels. Second, we examined the vanilla versions of these models without fine-tuning their parameters to human data, particularly in the cognitive IBL models, where an equal credit assignment mechanism leads to more conservative exploratory strategies. Future research could enhance these LLM-based and cognitive prediction models to better match human behavior and uncover new behavioral structures. Finally, while our study highlights the strengths of different models, it does not fully explore the integration of LLMs and cognitive models in a cohesive framework, which we consider as future work. Additionally, it would be useful to investigate the potential of these models in interactive systems to predict human decisions in real-time.

% \section{Conclusion}
% \label{sec:conclusion}
% \input{7.conclusion}
\section{Acknowledgements}
This research was supported by the University of Dayton Research Council Seed Grant and the UD/UDRI Summer Research Fellows Program.
\appendix
\section{Task Instructions}\label{sec:instructions}

The full task instruction provided to the model is:

\begin{quote}
\texttt{In a gridworld with obstacles represented by black blocks, a person navigates to find a goal with the highest score among four goals: blue, green, orange, and purple. Movement is restricted to up, down, left, and right directions within the grid. 
Each episode allows a maximum of 31 steps, with a total of 40 episodes permitted. The score is determined by reaching a target, with a penalty of 0.01 points for each step taken and 0.05 points for colliding with an obstacle. The objective is to locate the highest value target within the grid.}

\texttt{Given the current position at (x, y),
Moving up will result in the new position (x, y + 1),
Moving down will result in the new position (x, y - 1),
Moving right will result in the new position (x + 1, y),
Moving Left will result in the new position (x - 1, y).}
\end{quote}

\bibliography{references}

\end{document}